\def\year{2025}\relax
\documentclass[letterpaper]{article} % DO NOT CHANGE THIS
\usepackage{aaai2026}  
\usepackage{times}
\usepackage{helvet}
\usepackage{courier}
\usepackage[hyphens]{url}
\usepackage{graphicx}
\usepackage{natbib}
\usepackage{caption}
\usepackage{enumitem}
\usepackage{booktabs}
\usepackage{threeparttable}
\usepackage{subcaption}
\usepackage{amsmath}
\usepackage{amssymb}
\usepackage{float} 
\usepackage[none]{hyphenat}
\title{Exposing DeepFakes via Hyperspectral Domain Mapping}

\author {
    Aditya Mehta,
    Swarnim Chaudhary,
    Pratik Narang,
    Jagat Sesh Challa
}
\affiliations {
    Department of Computer Science and Information Systems,\\Birla Institute of Technology and Science, Pilani, Pilani Campus, Vidya Vihar, Pilani, Rajasthan 333031, India\\
    \{p20230303, f20231040, pratik.narang, jagatsesh\}@pilani.bits-pilani.ac.in\\
    }

\begin{document}
\maketitle
% \begin{figure*}[!b]
%   \centering
%   \includegraphics[width=0.95\textwidth]{Figures/intro figure.png}
%   \caption{Motivation: Hyperspectral imaging offers richer spectral information than RGB, revealing deepfake artifacts—such as distortions near the eyes—that remain hidden in RGB but become prominent at lower frequencies. }
%   \label{fig:intro}
% \end{figure*}

\begin{abstract}
Modern generative and diffusion models produce highly realistic images that can mislead human perception and even sophisticated automated detection systems. Most detection methods operate in RGB space and thus analyze only three spectral channels. We propose \textbf{HSI-Detect}, a two-stage pipeline that reconstructs a 31-channel hyperspectral image from a standard RGB input and performs detection in the hyperspectral domain. Expanding the input representation into denser spectral bands amplifies manipulation artifacts that are often weak or invisible in the RGB domain, particularly in specific frequency bands. We evaluate \textbf{HSI-Detect} across FaceForensics++ dataset and show the consistent improvements over RGB-only baselines, illustrating the promise of spectral-domain mapping for Deepfake detection.
\end{abstract}

\section{Introduction}
Rapid progress in generative adversarial networks (GANs) and diffusion models has made it increasingly easy to synthesize highly realistic human faces, voices, and videos. These so-called \textit{deepfakes} are no longer confined to entertainment and creative media; they pose risks of misinformation, impersonation, and even legal or political manipulation. Consequently, robust and generalizable detection methods have become an urgent research need. 

A key limitation of current deepfake detectors is their reliance on RGB images, which capture only three broad spectral channels. While suitable for visualization, RGB compresses much of the fine spectral information in natural images, causing subtle generative artifacts to be averaged out. These artifacts often lie in narrow spectral bands or specific frequency ranges, making RGB-based detectors vulnerable to missed detections and poor cross-dataset generalization.

\begin{figure*}[!t]
  \centering
  \includegraphics[width=0.95\textwidth]{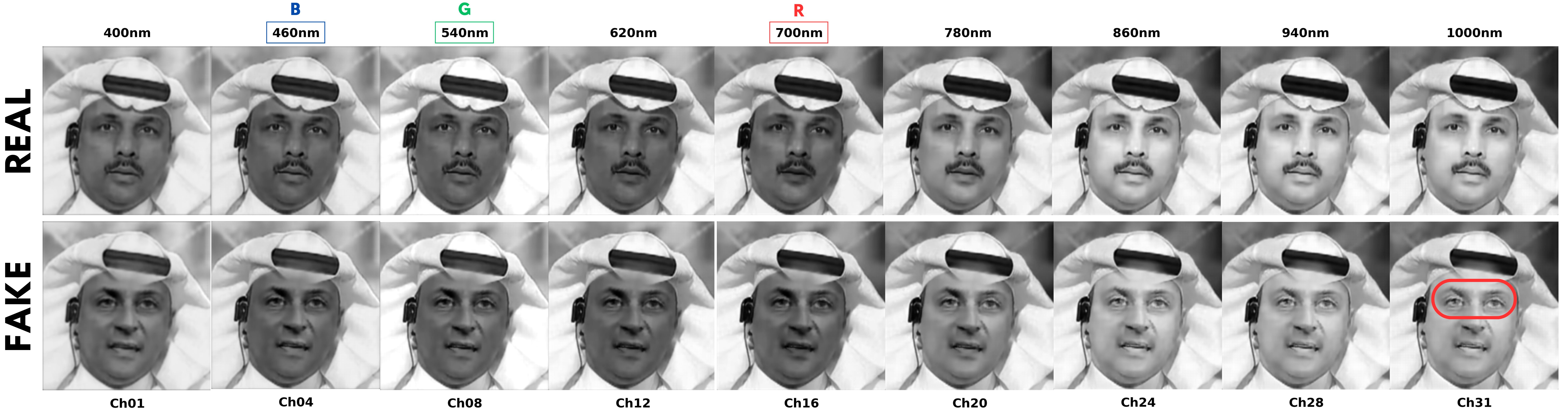}
  \caption{Motivation: Hyperspectral imaging offers richer spectral information than RGB, revealing deepfake artifacts—such as distortions near the eyes—that remain hidden in RGB but become prominent at lower frequencies.}
  \label{fig:intro}
\end{figure*}

Research has shown that spectral expansion can reveal inconsistencies invisible in RGB space, but most frequency-aware methods still operate on RGB inputs and remain limited by their three-channel representation. In contrast, hyperspectral imaging (HSI) captures tens or hundreds of narrow bands, enabling finer analysis of subtle variations. HSI has proven to be valuable in domains such as remote sensing \citep{peyghambari2021hyperspectral} and environmental monitoring \citep{wright2019raman}, where fine spectral cues are critical. Motivated by these findings, we propose that deepfake detection can similarly benefit from hyperspectral representations, as illustrated in Figure~\ref{fig:intro}.

% A key limitation of existing detectors is their reliance on standard RGB images, which provide only three spectral channels. RGB representations are efficient for visualization, but they compress much of the fine spectral information available in natural images. Subtle manipulation artifacts introduced by generative models are often spread across narrow spectral bands or manifest in frequency ranges that are averaged out in the RGB channels. This makes simple RGB-based detectors vulnerable to both missed detections and poor cross-dataset generalization.

% Previous research shows that spectral expansion can reveal latent inconsistencies that may not be visible in RGB domain. However, frequency-aware methods still operate on RGB inputs, and therefore remain constrained by the limited information captured by three broad bands. At the same time, hyperspectral imaging (HSI) has gained attention in remote sensing, medical imaging, and material analysis because it provides tens or even hundreds of narrow spectral bands, enabling more discriminative analysis of fine-grained signals. Inspired by these findings, we propose that deepfake detection can similarly benefit from hyperspectral representations.

In this work, we introduce \textbf{HSI-Detect}, a hyperspectral-guided deepfake detection framework. Instead of training detectors on RGB images alone, we reconstruct a 31-channel hyperspectral image from RGB inputs. These expanded spectral bands capture latent artifacts that generative models unintentionally introduce, particularly in low- and high-frequency regions \citep{dong2022think}. A dedicated classification network then analyzes the hyperspectral representation to detect whether the input is real or fake. By moving beyond the three-channel bottleneck, \textbf{HSI-Detect} seeks to improve robustness and generalization in the battle against deepfakes by providing a richer input to the detector.

\section{Proposed Method}

To propose HSI-Detect, we hypothesize that multi-band information in the reconstructed hyperspectral images can improve the separability between real and fake content by exposing spectral artifacts that remain hidden in RGB space. In particular, while RGB compresses the visible spectrum into three broad channels, hyperspectral reconstruction expands this into 31 narrow bands, allowing the detector to analyze subtle frequency irregularities and inter-band inconsistencies that generative models inadvertently introduce. Our method therefore consists of two main stages:

\begin{enumerate}[nosep,topsep=0pt]
\item \textbf{Hyperspectral Reconstruction (HSR).} To reconstruct hyperspectral images from RGB inputs, we use the MST++ model \citep{cai2022mst++}, a transformer-based framework tailored for spectral reconstruction. MST++ employs \textit{spectral-wise self-attention} to capture inter-band correlations often missed by CNNs focused on spatial features. Its multi-stage U-shaped encoder–decoder progressively refines outputs, enabling high-fidelity recovery of subtle spectral signatures. By emphasizing spectral self-similarity alongside local details, MST++ provides 31-channel hyperspectral estimates that form the input to our detection module.
  % \item \textbf{Hyperspectral Reconstruction (HSR).} To reconstruct hyperspectral images from RGB inputs, we adopt the MST++ model \citep{cai2022mst++}, a transformer-based framework specifically designed for spectral reconstruction. MST++ leverages \textit{spectral-wise self-attention} to capture correlations across spectral bands, a property that is often overlooked by CNN-based approaches focusing mainly on spatial features. Its multi-stage design, composed of U-shaped encoder–decoder blocks, progressively refines reconstructions from coarse to fine, enabling high-fidelity recovery of hyperspectral information from RGB inputs. Unlike conventional methods, MST++ emphasizes spectral self-similarity rather than purely spatial correlations, which makes it particularly effective at reconstructing subtle spectral signatures. This ability to capture both global spectral dependencies and local details makes MST++ an ideal backbone for our reconstruction stage, allowing HSI-Detect to generate 31-channel hyperspectral estimates that serve as input to the detection module.

  \item \textbf{Spectral Detection Network.} We employ an enhanced version of UCF \citep{yan2023ucf} to mitigate overfitting to specific forgery cues. The architecture uses a Disentanglement Framework that exploits spectral artefacts across 31 hyperspectral channels. It consists of an encoder, a decoder, and two classification heads. The encoder has a content encoder and a fingerprint encoder to extract content and forgery features, while the decoder applies Adaptive Instance Normalization (AdaIN) \citep{huang2017arbitrary} to reconstruct images from these features:  
\begin{equation}
\mathrm{AdaIN}(x, y) = \sigma(y) \left( \frac{x - \mu(x)}{\sigma(x)} \right) + \mu(y),
\end{equation}
where $\mathbf{x}$ is the content vector, $\mathbf{y}$ the style vector, and $\mu$, $\sigma$ denote mean and standard deviation.  
\end{enumerate}

To support detection, we introduce three loss functions: (1) Multi-task Classification loss for learning forgery-specific and shared features, (2) Contrastive Regularization loss to enhance discrimination between real and fake, and (3) Reconstruction loss to ensure consistency between original and reconstructed images.  

%   \item \textbf{Spectral Detection Network.} {We are using an enhanced version of UCF \citep{yan2023ucf} to address overfitting to specific forgery cues in deepfake detection. The proposed architecture consists of a Disentanglement Framework that exploits the spectral artefacts present in 31 channels of Hyperspectral Images. Our Disentanglement Framework consists of an encoder, a decoder and two classification heads. The encoder contains a content encoder and fingerprint encoder which extract content and forgery features respectively. The decoder uses Adaptive Instance Normalization (AdaIN) \citep{huang2017arbitrary} to reconstruct an image using the content and forgery features upsampling and convolutional operations. The formula is: 
%   \begin{equation}
% \mathrm{AdaIN}(x, y) = \sigma(y) \left( \frac{x - \mu(x)}{\sigma(x)} \right) + \mu(y),
% \end{equation}
% where $\mathbf{x}$ is the content vector and $\mathbf{y}$ is the style vector of an image. The functions $\mu(\cdot)$ and $\sigma(\cdot)$ represent the mean and standard deviation of the input, respectively.

% To aid the Detection Network, we introduce three loss functions: Multi-task Classification loss to guide the model to learn forgery-specific and common features, Contrastive Regularization loss to optimize the differentiating ability of model between fake and real images and Reconstruction loss to ensure the consistency of reconstructed and original image.}
% \end{enumerate}

\section{Experimental Evaluation}
Our model has been trained on Neural Textures, one of the four manipulation techniques in FaceForensics++ \citep{rossler2019faceforensics++} dataset. All preprocessing and training codebases utilized in this study follow DeepfakeBench \citep{yan2023deepfakebench} to ensure alignment with standardized settings. We utilize the Area Under the Curve (AUC) of the Receiver Operating Characteristic (ROC) as our evaluation metric. AUC quantifies the area beneath the ROC curve. A higher AUC score indicates better detection performance. For comparison with State-of-the-Art methods, we incorporate three baseline detectors: ViT {\citep{dosovitskiy2020image}}, RECCE {\citep{cao2022end}} and MoE-FFD {\citep{kong2025moe}}.

As shown in Table~\ref{table:results}, HSI-Detect consistently outperforms prior methods across multiple unseen manipulation types, achieving the best overall average AUC. The gains are especially clear on DeepFakes and FaceSwap, where hyperspectral cues provide strong discriminative power. These results highlight that our approach not only improves over RGB-only detectors but also advances beyond recent high-quality conference benchmarks, establishing HSI-Detect as a promising step forward for more robust and generalizable deepfake detection.

\begin{table}[t]
\centering
\begin{threeparttable}
\scalebox{0.88}{
\begin{tabular}{lcccc}
\toprule
\textbf{Method} & \textbf{DF} & \textbf{FF} & \textbf{FS} & \textbf{AVG}\\
\midrule
ViT (ICLR'21) & 78.46 & 68.31 & 45.07 & 63.95\\
RECCE (CVPR'22) & 72.37 & 64.69 & 51.61 & 62.89\\
MoE-FFD (TDSC'25) & 80.02 & \textbf{73.02} & 51.94 & 68.33\\
\textbf{HSI-Detect (Ours)} & \textbf{85.31} & 67.31 & \textbf{54.15} & \textbf{68.92} \\
\bottomrule
\end{tabular}
}
\end{threeparttable}
\caption{HSI-Detect achieves the highest average AUC for cross-manipulation detection, demonstrating the advantage of hyperspectral features over RGB-only baselines.}
\label{table:results}
\end{table}

\section{Conclusion and Future Work}
We introduce HSI-Detect, a framework that leverages reconstructed hyperspectral images for deepfake detection. Our experiments show that extending beyond RGB enables multi-band spectral cues to reveal artifacts often hidden in standard visual space, confirming the promise of hyperspectral analysis for forensics.

Future progress hinges on two directions: improving hyperspectral reconstruction, ideally with face-focused training to capture subtle details, and designing detection architectures tailored to hyperspectral inputs rather than adapted from RGB models.

Overall, HSI-Detect establishes a strong proof-of-concept and highlights the potential of hyperspectral imaging to deliver more robust and generalizable deepfake detection systems.
\bibliography{aaai2026}
\end{document}